\documentclass[11pt, a4paper, logo, copyright, nonumbering]{deepseek}
\usepackage[numbers, square, sort&compress]{natbib}
\usepackage{dblfloatfix}
\usepackage{ulem}
\usepackage{caption}
\definecolor{citecolor}{HTML}{1976D2}
\hypersetup{
    colorlinks=true,
    linkcolor=black,
    filecolor=magenta,
    urlcolor=cyan,
    citecolor=citecolor,
}
\usepackage{dramatist}
\usepackage{xspace}
\usepackage{pifont}
\usepackage{multirow}
\usepackage{tcolorbox}
\usepackage{xltabular}

\usepackage{longtable}
\usepackage{hyperref}
\usepackage{diagbox}
\usepackage{makecell}
\usepackage{amsfonts}
\usepackage{amsmath}
\usepackage{amssymb}
\usepackage{lineno}

\usepackage[bottom]{footmisc}

\usepackage{CJKutf8}
\usepackage{subfigure}
\usepackage{setspace}
\usepackage{subcaption} 
\usepackage{epigraph}
\usepackage{wrapfig}
\usepackage{algorithm}
\usepackage{algpseudocode}
\usepackage{amsmath}

\makeatletter
\def\@BTrule[#1]{%
  \ifx\longtable\undefined
    \let\@BTswitch\@BTnormal
  \else\ifx\hline\LT@hline
    \nobreak
    \let\@BTswitch\@BLTrule
  \else
     \let\@BTswitch\@BTnormal
  \fi\fi
  \global\@thisrulewidth=#1\relax
  \ifnum\@thisruleclass=\tw@\vskip\@aboverulesep\else
  \ifnum\@lastruleclass=\z@\vskip\@aboverulesep\else
  \ifnum\@lastruleclass=\@ne\vskip\doublerulesep\fi\fi\fi
  \@BTswitch}
\makeatother

\addto\extrasenglish{
}

 {\begin{list}{}%
         {\setlength{\leftmargin}{#1}}%
         \item[]%
 }
 {\end{list}}
 
\reportnumber{001} 

\renewcommand{\today}{}

\title{\centering
    Unlimited OCR Works 2\\
    {\large\normalfont\textit{Pioneering the OCR-3.0 Era: From Long- to Unlimited-Horizon Parsing}}
}
\vspace{-1cm}

\author{
Baidu ERNIE Team\\
}

\renewcommand{\phi}{\varphi}

\renewcommand{\epsilon}{\varepsilon}
\renewcommand{\imath}{\mathrm{i}}

\newlength{\restsubwidth}
\newlength{\restsubheight}
\newlength{\restsubmoreheight}
\newcommand{\rest}[2]{%
        \settowidth{\restsubwidth}{\ensuremath{#2}}
        \settoheight{\restsubheight}{\ensuremath{{}_{#2}}}
        \ensuremath{{#1\hskip 0.5pt}_{\vrule\kern2pt\parbox[b][%
        4pt][b]{\the\restsubwidth}{%
                        \ensuremath{{}_{#2}}}}}
        }

\vspace{-1cm}
\begin{abstract}

\vspace{-1.5em}
Unlimited OCR ushers the era of long-horizon parsing, substantially mitigating the growth of working memory as document lengthens. However, as document length scales to hundreds of pages, the visual context grows substantially, sharply increasing the computational cost of each decoding step and limiting the scalability of ultra-long document parsing. Inspired by human book transcription via localized focus without re-scanning preceding pages, we introduce Unlimited OCR 2, a document parsing model that looks only at what it transcribes. Building on Unlimited OCR, we upgrade all the decoder attention modules to Sparse Reference Sliding Window Attention (SR-SWA), bounding active reference tokens to a fixed capacity and thus sustaining constant computational costs across arbitrarily long documents. By combining SR-SWA with Unlimited-Horizon Decoding Paradigm, Unlimited OCR 2 enables document parsing over virtually unlimited lengths without prohibitive memory overhead. At the hundred-page scale, Unlimited OCR 2 achieves 7.0$\times$ and 5.6$\times$ per-step speedups over Full Attention and R-SWA, respectively. Furthermore, it delivers state-of-the-art end-to-end performance on both MPDocBench and MarathonDocBench, our newly proposed benchmark targeting hundred-page document parsing. Ultimately, the capacity for uninterrupted, one-shot document parsing signifies a fundamental paradigm shift into the OCR 3.0 era.
\end{abstract}

\begin{document}

\begin{CJK*}{UTF8}{gbsn}

\maketitle
\vspace{-0.2cm}

\begin{figure}[h]
    \centering
    \vspace{-0.1cm}
    \includegraphics[width=0.8\linewidth]{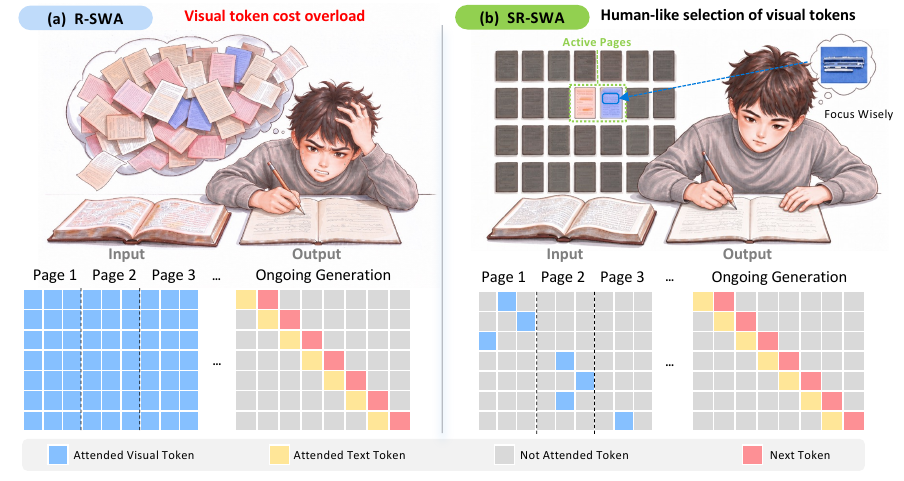}
    \vspace{-0.2cm}
    \caption{R-SWA vs SR-SWA. R-SWA computes attention over all visual tokens across pages, inducing extreme visual redundancy; SR-SWA sparsely attends to relevant visual regions, aligning machine decoding with human reading intuition.}
    
    \label{fig:1}
\end{figure}

\newpage

\begin{spacing}{0.9}
\tableofcontents
\end{spacing}

\newpage

\section{Introduction}

\begin{quote}
    \itshape
    “Pixels are my body, and tokens are my blood. I have parsed over a thousand pages. Unknown to KV cache. Nor known to context limits. So, as I pray: Unlimited OCR Works.”
    \hrule
    \vspace{-0.5em}
    \begin{flushright}
        \normalfont\footnotesize Snarcy on huggingface ( Adapted from \textit{UBW})
    \end{flushright}
\end{quote}
 \vspace{-0.5em}
\vspace{-0.2cm}

Document parsing has undergone a major paradigm shift, transitioning from traditional multi-stage pipelines (OCR-1.0)~\cite{cui2025paddleocr,zhang2026paddleocr,cui2026paddleocr,MinerU_2_5_2025,wang2026mineru2,MonkeyOCR_2025,duan2026glm} with handcrafted detection and recognition modules to end-to-end Vision-Language Models (OCR-2.0)~\cite{li2025dots,dong2026qianfan,lu2026ovisocr2,wu2026firered,li2026hunyuanocr}. By formulating document understanding as a unified sequence-to-sequence transcription task, OCR-2.0 frameworks effectively eliminate cumulative pipeline errors. Our previous work, Unlimited OCR~\cite{yin2026unlimited}, pushed OCR-2.0 boundaries to tens of pages by using Reference Sliding Window Attention (R-SWA) to maintain a constant KV cache. Nevertheless, it still computes attention over all visual tokens across pages during decoding. As document length scales, this persistent attention to non-informative visual tokens severely escalates visual redundancy, compounding the difficulty of end-to-end inference. This visual attention bottleneck naturally prompts a fundamental question: how can we transition from length-limited parsing to an unlimited one-shot paradigm and genuinely inaugurate the OCR-3.0 era?

To answer this question, human transcription of lengthy books provides a clear physical intuition for parsing scaling: as we copy text page by page, we maintain localized visual focus, casting only momentary glances back when cross-page references are needed, rather than keeping the entire volume active in mind. This cognitive mechanism reveals that visual dependencies in multi-page documents are inherently page-sparse yet intra-page concentrated. However, standard dense global attention completely overlooks this sparse dynamic, incurring redundant attention interactions across all pages. This not only incurs prohibitive computational overhead, but also leads to severe memory bottlenecks as document length scales. To break through these efficiency and memory bottlenecks, we align machine perception with human reading patterns and present Unlimited OCR 2, pioneering an unlimited-horizon parsing framework for the OCR-3.0 era. Our main contributions are as follows:
\vspace{-0.2cm}
\begin{itemize}
    \item We introduce Sparse Reference Sliding Window Attention (SR-SWA), a hierarchical sparse attention mechanism designed for unlimited visual context. Specifically, Page Gate first narrows each query's visual reference scope from the entire document to relevant pages while robustly modeling multi-page positional relationships. Token Sieve selectively incorporates top-$k$ remote token buckets to capture long-range visual dependencies, thereby drastically reducing attention overhead while preserving comprehensive context.
    
   \item We introduce MarathonDocBench, a benchmark for hundred-page document parsing. It substantially extends existing evaluations in document length and cross-page structural complexity, enabling a comprehensive assessment of long-document parsing performance in the hundred-page regime.
    \item Comprehensive evaluations in long-horizon settings demonstrate that our approach maintains constant per-step decoding latency regardless of document length while achieving state-of-the-art end-to-end performance on multi-page benchmarks. Building on these advances, we establish an efficient unlimited-horizon parsing paradigm by integrating SR-SWA with the Unlimited-Horizon Decoding Paradigm, enabling document parsing over virtually unlimited lengths.
\end{itemize}
\vspace{-0.2cm}

In summary, SR-SWA fundamentally breaks the computational and memory bottlenecks of ultra-long visual context. Operating at constant per-step decoding latency with length-invariant accuracy across hundreds of pages, Unlimited OCR 2 unlocks scalable long-document parsing and paves the way for the OCR-3.0 era of unlimited-horizon understanding.


\section{Related Works}
\label{sec:related1}


\subsection{VLM-based Document Parsing}
Document parsing has increasingly been reshaped by advances in vision-language models, evolving from pipeline-based OCR-1.0 systems~\cite{cui2025paddleocr,zhang2026paddleocr,cui2026paddleocr,MinerU_2_5_2025,wang2026mineru2,MonkeyOCR_2025,duan2026glm}, which combine specialized page analysis modules with VLM-based decoding, to unified VLM-based OCR-2.0 systems that directly generate structured representations from document images. OCR-2.0 encompasses both single-page parsers~\cite{li2025dots,dong2026qianfan,wei2026hpd,lu2026ovisocr2,wu2026firered,li2026hunyuanocr} and Unlimited OCR~\cite{yin2026unlimited}, which extends end-to-end parsing to multi-page documents by jointly modeling their visual content within a shared context. However, although Unlimited OCR improves long-document efficiency with R-SWA, it still densely attends to all remote visual tokens, causing the visual context accessible to each decoding query to expand with document length and creating a fundamental scalability bottleneck at the hundred-page scale. Unlimited OCR 2 overcomes this limitation by explicitly bounding remote visual access, enabling attention-efficient end-to-end parsing of hundred-page documents and advancing document parsing into the OCR-3.0 era.

\subsection{Efficient Long-Context Modeling}
Efficient long-context modeling has been explored through both attention replacement and sparse attention. Linear or recurrent alternatives~\cite{gu2023mamba,yang2023gated,katharopoulos2020transformers,yang2024parallelizing,li2025minimax} reduce long-context cost by summarizing historical information into compact states. Document parsing, however, requires long-range memory to preserve fine-grained visual evidence for accurate recognition and cross-page retrieval, which may not be fully retained within such compact representations. 

Sparse attention retains direct access to selected historical tokens or KV blocks while avoiding dense all-to-all computation. Early methods use predefined patterns such as local windows and global tokens~\cite{beltagy2020longformer,zaheer2020big}, whereas recent approaches adopt query-dependent retrieval. Hard-selection methods, such as  Compressed Sparse Attention(CSA)~\cite{xu2026deepseek} and MiniMax Sparse Attention(MSA)~\cite{lai2026minimax}, rely on discrete top-$k$ routing and typically require auxiliary supervision or warmup. In contrast, DashAttention~\cite{huang2026dashattention} and HiLS~\cite{hu2026hierarchical} incorporate differentiable or forward-integrated routing scores, enabling more direct end-to-end optimization.

\section{Methodology}
\subsection{Problem Setup}

\subsubsection{Unlimited-horizon Parsing}
Unlimited-horizon parsing aims to enable end-to-end processing of documents of arbitrary lengths, from single-page forms to massive multi-page documents, with bounded computational and memory complexities.
Formally, multi-page document parsing generates a transcription $\mathcal{Y}=(y_1,\ldots,y_L)$ from an ordered document $\mathcal{D}=(P_1,\ldots,P_N)$ with visual tokens $\mathcal{V}_{1:N}$. At decoding step $t$, the model operates on:
\begin{equation}
    \mathcal{X}_t = \operatorname{Concat}\left(\mathcal{V}_1,\ldots,\mathcal{V}_N, y_{<t}\right).
\end{equation}
While Unlimited OCR constrained text history to a fixed sliding window via R-SWA, it still requires attending over all preceding visual tokens $\mathcal{V}_{1:N}$. In the hundred-page regime, this page-dependent visual expansion becomes the primary bottleneck. Decoupling visual context from page count \(N\) to maintain a bounded computational and memory footprint directly motivates our proposed Sparse Reference Sliding Window Attention (SR-SWA).

\begin{figure}[t]
    \centering
    \includegraphics[width=1\linewidth]{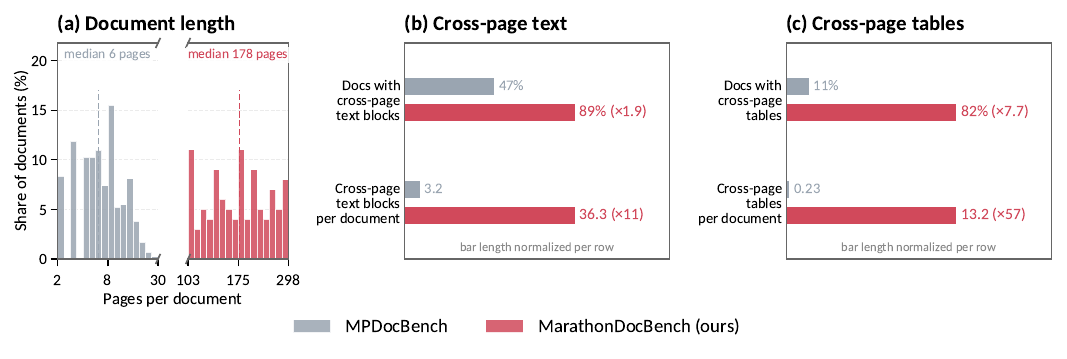}
    \vspace{-0.8cm}
    \caption{MarathonDocBench versus MPDocBench. MarathonDocBench extends document length from tens to hundreds of pages (a), with substantially richer cross-page structures as shown in (b) and (c).}
    \label{fig:dataset}
    \vspace{-0.2cm}
\end{figure}

\subsubsection{MarathonDocBench}

Existing multi-page document parsing benchmarks neither extend to the hundred-page scale nor sufficiently capture the rich cross-page structures that emerge in long documents. To address this gap, we introduce MarathonDocBench, the first benchmark specifically designed for hundred-page document parsing. It contains 100 documents spanning 103--300 pages with 205k annotated layout elements, extending the document-length range by roughly an order of magnitude beyond MPDocBench~\cite{zhou2026mpdocbench}.

Beyond document length, MarathonDocBench contains substantially richer cross-page structures than MPDocBench, with cross-page text and tables appearing much more frequently and densely across documents, as summarized in Figure~\ref{fig:dataset}. This makes MarathonDocBench a substantially more challenging testbed for evaluating whether end-to-end parsers can preserve document-level structure while scaling from tens to hundreds of pages. To ensure reliable evaluation, structural annotations, including bounding boxes, block type, hierarchy, reading order, and cross-page continuations of text and tables, are initialized from a strong page-by-page parser and carefully refined by human annotators. Content transcriptions are initialized from the same parser outputs and verified through an iterative VLM-based refinement pipeline, with flagged disagreements adjudicated by human annotators.


\subsection{Architecture}
\begin{figure}[t]
	\centering
    \includegraphics[width=1\linewidth]{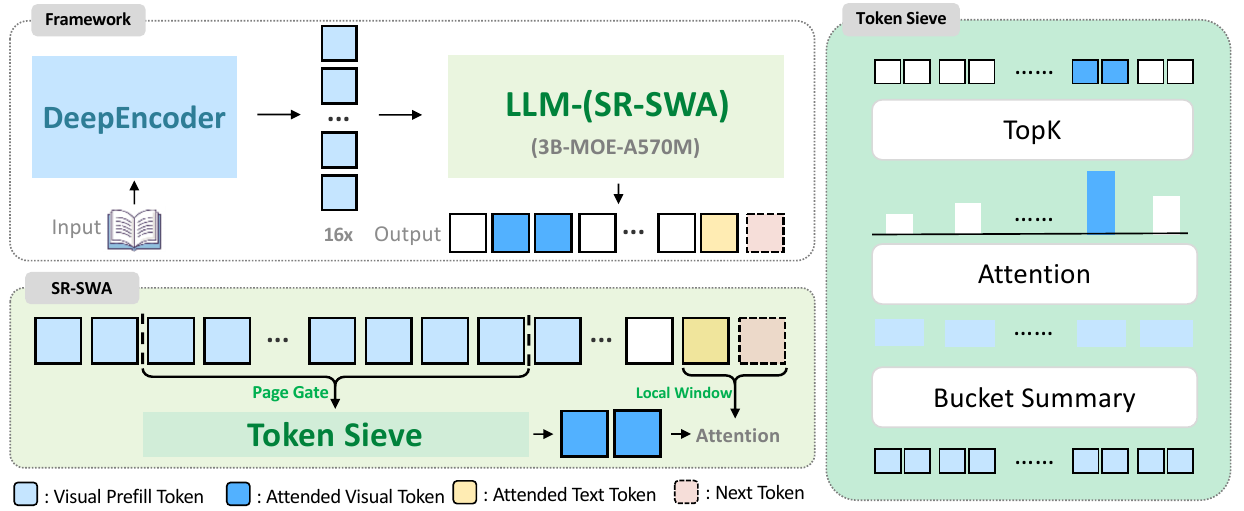}
	\caption{Inspired by the process of humans copying books, we propose Unlimited OCR 2. SR-SWA combines Page Gate and Token Sieve to construct a fixed-length set of working reference visual tokens, substantially reducing computational and memory overhead while keeping both independent of document length throughout generation.}
	\label{fig:architecture}
\end{figure}
Unlimited OCR 2 inherits the fundamental backbone from Unlimited OCR, leveraging DeepSeek OCR~\cite{wei2025deepseek} as its foundation. Specifically, it employs the DeepEncoder paired with a Mixture-of-Experts (MoE) decoder architecture featuring 3B total parameters and 500M activated parameters per token. Building upon this base, we replace the baseline R-SWA with our proposed SR-SWA as shown in Figure~\ref{fig:architecture}. While R-SWA bounds local interactions, its visual context scope still expands with growing document length, leading to noticeable latency accumulation over ultra-long horizons. By introducing a selective remote visual routing mechanism, SR-SWA strictly bounds the active context scope, enabling Unlimited OCR 2 to maintain a constant KV cache footprint across both text and visual modalities during unlimited-horizon decoding.  We detail the inner mechanics of SR-SWA in the following subsections.

\subsection{Sparse Reference Sliding Window Attention}

Following standard block-sparse attention~\cite{lai2026minimax,huang2026dashattention,xu2026deepseek,xu2026deepseek_flash},
we partition $\mathcal{X}_t$ into fixed-size blocks $\{\mathcal{B}_b\}_{b=1}^{B}$
and preserve a fixed local window $\mathcal{L}(q)$ for each query $q$.
SR-SWA constructs $\mathcal{S}(q)$ hierarchically,
using Page Gate for page-level organization and Token Sieve for fixed-budget selection.

\subsubsection{Page Gate}
To eliminate irrelevant visual tokens and concentrate focus on the current page, Page Gate restricts the visual attention scope for a query $q$ associated with page $g(q)$ by retaining only blocks from designated target pages:
\begin{equation}
    \mathcal{C}_{\mathrm{page}}(q) = \left\{ b \;\middle|\; \mathcal{B}_b \cap \mathcal{V}_{g(q)} \neq \emptyset \right\}.
\end{equation}
Since block boundaries are independent of page boundaries, a candidate block may straddle adjacent pages. Meanwhile, to prevent positional explosion as page count scales, Page Gate factorizes each token position into a bounded page-level coordinate (following an alternating $0$--$1$ pattern) and a monotonic intra-page coordinate:
\begin{equation}
    \mathbf{p}_{i,j} = \left[ \mathbf{p}^{\mathrm{page}}_i,\; \mathbf{p}^{\mathrm{intra}}_{i,j} \right],
\end{equation}
ensuring both stable length extrapolation and efficient cross-page visual routing.
\subsubsection{Token Sieve}
While Page Gate restricts the routing space across pages, efficient fixed-budget routing requires a principled block selection strategy. Conventional sparse routers typically rely on dense-attention distillation, which lacks direct alignment with the language modeling objective. Inspired by recent end-to-end routing designs~\cite{huang2026dashattention,hu2026hierarchical}, Token Sieve directly optimizes fixed-budget block selection under the language modeling objective.

Given the candidate set $\mathcal{C}_{\mathrm{page}}(q)$ from Page Gate, Token Sieve summarizes each candidate block $b$ into a compact key representation $\bar{\mathbf{k}}_b$ and computes its relevance to query $q$. The router then forms a sparse routing distribution and retains only the top-$k$ blocks:
\begin{equation}
\begin{aligned}
s_b(q)
&=
\frac{\tilde{\mathbf{q}}^\top \bar{\mathbf{k}}_b}{\sqrt{d}},
\qquad
\boldsymbol{\pi}(q)
=
\alpha\text{-entmax}
\left(
\{s_b(q)\}_{b\in\mathcal{C}_{\mathrm{page}}(q)}
\right), \\
\mathcal{S}(q)
&=
\operatorname{TopK}
\big(
\boldsymbol{\pi}(q), k
\big).
\end{aligned}
\end{equation}
Here, $d$ denotes the hidden dimension. By selectively attending to a fixed number of top-$k$ remote buckets, Token Sieve captures long-range visual dependencies with minimal computational overhead. To preserve relative preferences among selected blocks, the routing probabilities are converted into a mean-centered log-prior:
\begin{equation}
    \beta_b(q)
    =
    \log \pi_b(q)
    -
    \frac{1}{|\mathcal{S}(q)|}
    \sum_{b' \in \mathcal{S}(q)}
    \log \pi_{b'}(q),
    \qquad b \in \mathcal{S}(q).
\end{equation}
\vspace{-0.2cm}

With the selected remote blocks $\mathcal{S}(q)$ and the fixed local window $\mathcal{L}(q)$, each query attends to a length-agnostic retained token set defined as:
\begin{equation}
    \mathcal{R}(q)
    =
    \mathcal{L}(q)
    \cup
    \bigcup_{b\in\mathcal{S}(q)}
    \mathcal{B}_b .
    \vspace{-0.2cm}
\end{equation}
The attention weights over $\mathcal{R}(q)$ are computed as:
\begin{equation}
    \alpha_j(q)
    =
    \operatorname{Softmax}
    \left(
        \frac{q^\top k_j}{\sqrt{d}}
        +
        \mathbb{I}\!\left[b(j)\in\mathcal{S}(q)\right]
        \beta_{b(j)}(q)
    \right),
\end{equation}
where the softmax is normalized over all tokens in $\mathcal{R}(q)$, $b(j)$ denotes the block containing token $j$, and $k_j$ denotes its key projection. Tokens within the local window use standard attention logits, while tokens from selected remote blocks additionally incorporate the corresponding routing bias. Since the routing bias directly participates in these logits, gradients from the parsing objective can propagate to the routing module, enabling end-to-end optimization without an auxiliary KL-divergence distillation loss.

\vspace{-0.2cm}

\begin{algorithm}[!h]
\caption{Cross-Chunk KV Reuse for Unlimited-Horizon Decoding}
\label{alg:page_split}
\begin{algorithmic}[1]
\Require Pages $P_1,\dots,P_N$; instruction $q$; chunk size $W$; tail length $T$
\Ensure Transcription $Y$
\Statex \textit{Notation:} $\mathcal{C}$ is a KV cache; $\oplus$ concatenates KV in order;
\Statex \hspace{\algorithmicindent} $\textsc{Enc}(X \mid \mathcal{C})$ returns the KV of input $X$ computed with $\mathcal{C}$ as context.
\State Partition $N$ pages into $K$ overlapping chunks $C_1,\dots,C_K$ with a 1-page boundary overlap \Comment{$K{=}1$ indicates unsegmented}
\State $\mathcal{C} \gets \textsc{Enc}(C_1 \oplus q)$ \Comment{only the first chunk carries $q$}
\State $y_1 \gets \textsc{Decode}(\mathcal{C})$;\quad $Y \gets y_1$
\For{$k = 2$ \textbf{to} $K$}
    \State $\mathcal{C} \gets \mathcal{C}^{\mathrm{img}}_{\text{overlap}} \oplus \textsc{Enc}(C_k \mid \mathcal{C}^{\mathrm{img}}_{\text{overlap}}) \oplus \mathcal{C}^{\mathrm{txt}}_{\text{tail}}$
    \State $y_k \gets \textsc{Decode}(\mathcal{C})$;\quad $Y \gets Y \oplus y_k$
\EndFor
\State \Return $Y$
\end{algorithmic}
\end{algorithm}

\begin{figure}[t]
	\centering
    \includegraphics[width=1\linewidth]{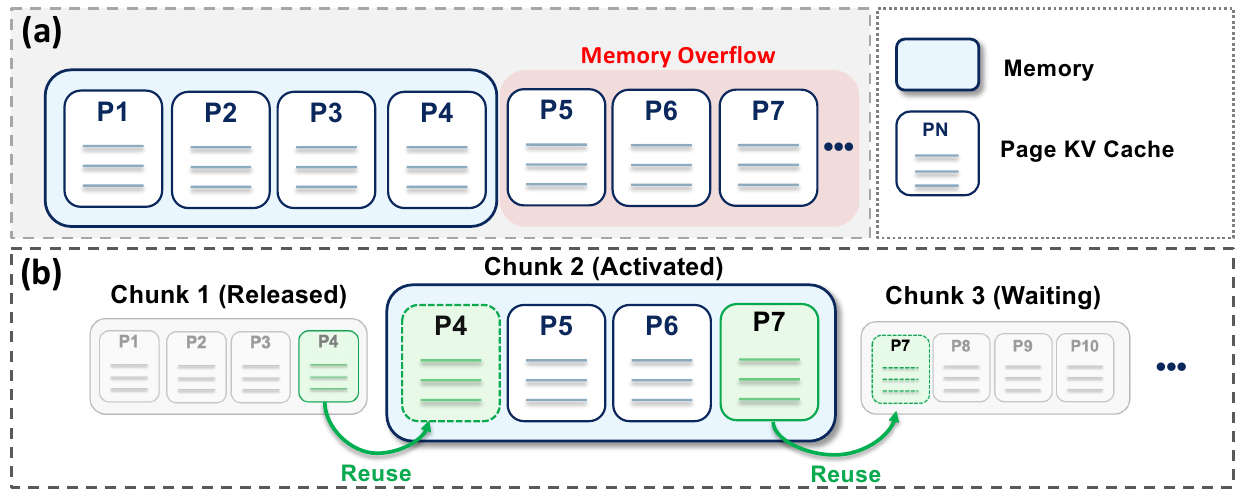}
    \vspace{-0.7cm}
	\caption{Comparison of (a) standard decoding and (b) our Unlimited-Horizon Decoding paradigm. Standard decoding retains visual KV caches for all pages, causing GPU memory usage to grow linearly with document length and eventually overflow. Our paradigm divides long documents into manageable chunks and processes them sequentially, reusing the KV cache of shared boundary pages and releasing inactive caches to keep GPU memory usage bounded while preserving cross-chunk continuity.}
    \vspace{-0.6cm}
	\label{fig:unlimited_horizon_decoding}
\end{figure}
\vspace{-0.6cm}

\subsection{Unlimited-Horizon Decoding Paradigm}

While SR-SWA bounds the number of reference tokens accessed by each query, processing extremely long multi-page documents remains constrained by the GPU memory required to prefill and retain their visual KV caches. As illustrated in Figure~\ref{fig:unlimited_horizon_decoding}, conventional all-at-once prefill keeps all page-level visual KV caches in memory, causing the memory footprint to grow linearly with document length. To overcome this limitation, we introduce the Unlimited-Horizon Decoding Paradigm, which partitions the document into overlapping page-level chunks and processes them sequentially. Adjacent chunks share their boundary page, whose KV cache is directly reused to preserve cross-chunk continuity, while visual KV caches outside the active chunk are released. Consequently, only a fixed-size chunk remains resident in GPU memory, bounding the memory footprint independently of the total document length. The complete procedure is formally described in Algorithm~\ref{alg:page_split}.

\vspace{-0.3cm}

\section{Experimental Settings}
\subsection{Data Engine}
\label{sec:data_engine}
To train Unlimited OCR 2 for versatile document parsing, we construct a large-scale multimodal corpus comprising $2\text{M}$ multi-page and $10\text{M}$ single-page documents. Pseudo-labels across both single- and multi-page subsets are extracted from MinerU2.5-Pro~\cite{wang2026mineru2} and PaddleOCR-VL-v1.6~\cite{zhang2026paddleocr}, tailored to their respective domain strengths. To enrich visual and layout diversity, we additionally incorporate versatile synthetic document generation pipelines. Across both real and synthetic data, we leverage VLM-based multi-dimensional feature analysis to conduct balance filtering, ensuring uniform coverage over complex layout distributions. Finally, to explicitly fortify structural hierarchy and cross-page continuity in multi-page scenarios, we apply MinerU-Popo~\cite{xu2026mineru} refinement alongside dedicated heuristic post-processing rules to reconcile layout inconsistencies across adjacent page boundaries.
\subsection{Implementation Details}
We initialize Unlimited OCR 2 from the Unlimited OCR checkpoint and perform post-training for 1 epoch on the curated dataset. Training is conducted on a cluster of $25 \times 8$ GPU NVIDIA H800 nodes with a global batch size of 1,600 and a maximum sequence length of 64K tokens. During training, we freeze both the DeepEncoder and the aligner, optimizing only the LLM parameters. Optimization is carried out using the AdamW optimizer with a cosine learning rate decay schedule, starting at an initial learning rate of $2.5 \times 10^{-4}$. The entire training workflow is implemented using the MS-Swift framework. For deployment, the Unlimited-Horizon Decoding Paradigm is integrated into SGLang to facilitate efficient long-document inference. 


\section{Evaluation}

\subsection{Benchmarks and Metrics}
\begin{figure}[b!]
	\centering
    \includegraphics[width=1\linewidth]{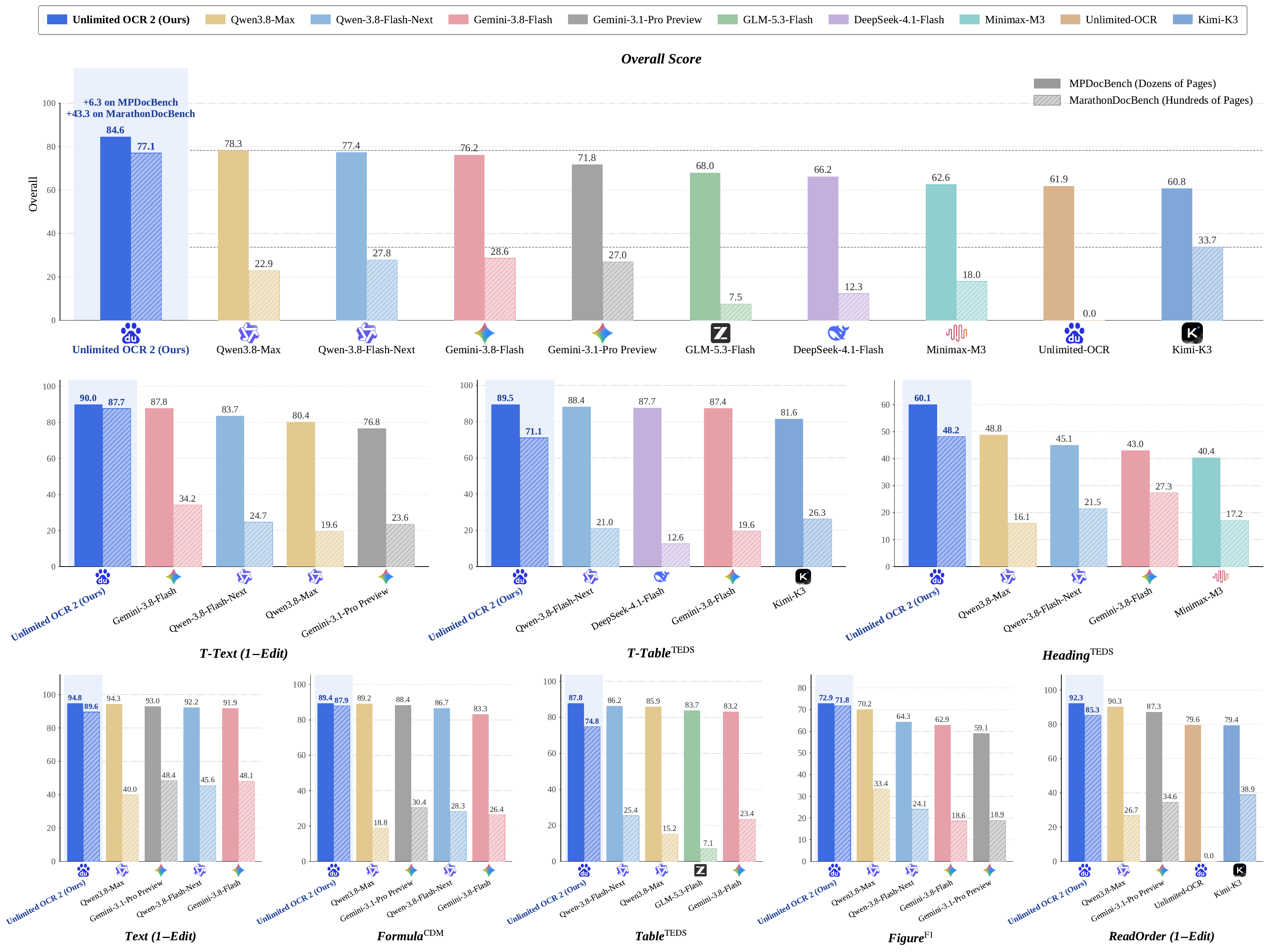}
	\caption{Performance comparison on multi-page benchmarks. Unlimited OCR 2 (Ours) consistently achieves state-of-the-art performance across all evaluation categories.}
	\label{fig:leaderboard}
\end{figure}
\label{sec:benchmarks_and_metrics}

\begin{table}[t]
\footnotesize
\caption{
Performance comparison on MPDocBench. All models in the table are end-to-end
VLM-based architectures.
}
\vspace{-0.2cm}
\label{tab:mpd}
\footnotesize
\setlength{\tabcolsep}{2.8pt}
\renewcommand{\arraystretch}{1.5}
\resizebox{\textwidth}{!}{
\begin{tabular}{
l|
c|
c
>{\columncolor{blue!5}}c
c
c
>{\columncolor{blue!5}}c
c
c
>{\columncolor{blue!5}}c
}
\toprule
\textbf{Method}
& \textbf{Overall}$\uparrow$
& \textbf{Text}$^{\mathrm{Edit}}\downarrow$
& \textbf{T-Text}$^{\mathrm{Edit}}\downarrow$
& \textbf{Formula}$^{\mathrm{CDM}}\uparrow$
& \textbf{Table}$^{\mathrm{TEDS}}\uparrow$
& \textbf{T-Table}$^{\mathrm{TEDS}}\uparrow$
& \textbf{Figure}$^{\mathrm{F1}}\uparrow$
& \textbf{ReadOrder}$^{\mathrm{Edit}}\downarrow$
& \textbf{Heading}$^{\mathrm{TEDS}}\uparrow$
\\
\midrule
\rowcolor{gray!10}
\multicolumn{10}{c}{\textit{General VLMs}} \\
Gemini-3.1-Pro Preview~\cite{Gemini_3_1_pro}
& 71.79 &  0.070 &  0.232 & 88.37 &  81.68  & 61.26 & 59.08  &  0.127 &  26.94 \\
Qwen3.8-Max~\cite{tongyi} & 
78.30	& 0.057 & 	0.196 & 	{89.24} & {85.89} & 67.18 & 70.19 & {0.097} & 48.80\\
Kimi-K3~\cite{team2026kimi}
& 60.77 & 0.162 & 0.551 & 41.16 & 75.45 & 81.57 & 40.47  & 0.206 & 39.30 \\
Qwen3.6-plus~\cite{tongyi} & 72.14 & 0.096 & 0.245 & {88.77} & 82.99 & 60.50 & 64.80 & 0.179 & 32.06 \\
Qwen3-VL-235B~\cite{bai2025qwen3} & 73.98 & 0.087 & 0.187 & 84.71 & 81.35 & 61.25 & 63.25 & 0.136 & 42.34 \\
InternVL-3.5-38B~\cite{wang2025internvl3} & 57.01 & 0.133 & 0.515 & 84.30 & 69.86 & 51.69 & 7.97 & 0.199 & 27.01 \\
Minimax-M3~\cite{lai2026minimax} & 62.65 & 0.167 & 0.422 & 81.30 & 77.91 & 76.51 & 7.18 & 0.231 & 40.39 \\
Gemini-3.8-Flash~\cite{Gemini_3_8_flash} & 76.25 & 0.081 & 0.122 & 83.27 & 83.15 & 87.41 & 62.86 & 0.294 & 43.05 \\
Qwen-3.8-Flash-Next~\cite{tongyi} & 77.36 & 0.078 & 0.163 & 86.70 & 86.24 & 88.36 & 64.31 & 0.277 & 45.09 \\
DeepSeek-4.1-Flash~\cite{xu2026deepseek_flash} & 66.25 & 0.114 & 0.451 & 82.12 & 80.14 & 87.72 & 35.55 & 0.312 & 32.10 \\
GLM-5.3-Flash~\cite{zai2026glm53flash} & 68.02 & 0.111 & 0.344 & 81.86 & 83.70 & 70.71 & 43.09 & 0.288 & 39.09 \\
\midrule
\rowcolor{gray!10}
\multicolumn{10}{c}{\textit{Specialized VLMs (OCR-2.0)}} \\
Unlimited-OCR~\cite{yin2026unlimited}
& 61.89 & 0.167 & 0.605 & 73.71 & 73.51 & 60.76 & 53.19 & 0.204 & 31.58 \\
\midrule
\rowcolor{gray!10}
\multicolumn{10}{c}{\textit{Specialized VLMs (OCR-3.0)}} \\
\textbf{Unlimited-OCR 2 (Ours)}
& \textbf{84.59} & \textbf{0.052} & \textbf{0.100} & \textbf{89.39} & \textbf{87.78} & \textbf{89.54} & \textbf{72.91} & \textbf{0.077} & \textbf{60.10} \\

\bottomrule
\end{tabular}
}
\vspace{-0.2cm}

\end{table}

\begin{table*}[h!]
\centering
\caption{
Performance comparison on MarathonDocBench. Unlimited OCR is not reported on MarathonDocBench due to its 40-page input limitation.
}
\vspace{-0.2cm}
\label{tab:marathon}

\footnotesize
\setlength{\tabcolsep}{2.8pt}
\renewcommand{\arraystretch}{1.5}

\resizebox{\textwidth}{!}{
\begin{tabular}{
l|
c|
c
>{\columncolor{blue!5}}c
c
c
>{\columncolor{blue!5}}c
c
c
>{\columncolor{blue!5}}c
}
\toprule
\textbf{Method}
& \textbf{Overall}$\uparrow$
& \textbf{Text}$^{\mathrm{Edit}}\downarrow$
& \textbf{T-Text}$^{\mathrm{Edit}}\downarrow$
& \textbf{Formula}$^{\mathrm{CDM}}\uparrow$
& \textbf{Table}$^{\mathrm{TEDS}}\uparrow$
& \textbf{T-Table}$^{\mathrm{TEDS}}\uparrow$
& \textbf{Figure}$^{\mathrm{F1}}\uparrow$
& \textbf{ReadOrder}$^{\mathrm{Edit}}\downarrow$
& \textbf{Heading}$^{\mathrm{TEDS}}\uparrow$
\\
\midrule
\rowcolor{gray!10}
\multicolumn{10}{c}{\textit{General VLMs}} \\
Gemini-3.1-Pro Preview~\cite{Gemini_3_1_pro}
  & 26.97 & 0.516 & 0.764 & 30.44 & 28.48 & 21.71 & 18.94 & 0.654 & 9.68 \\
Qwen3.8-Max~\cite{tongyi}
 & 22.93 & 0.600 & 0.804 & 18.82 & 15.20 & 13.71 & 33.39 & 0.733 & 16.09 \\
Kimi-K3~\cite{team2026kimi}
  & 33.74 & 0.475 & 0.664 & 40.50 & 27.80 & 26.30 & 21.38 & 0.611 & 28.90 \\
Minimax-M3~\cite{lai2026minimax}
 & 17.97 & 0.659 & 0.818 & 14.20 & 17.79 & 14.84 & 3.67 & 0.763 & 17.16 \\
Gemini-3.8-Flash~\cite{Gemini_3_8_flash}
  & 28.61 & 0.519 & 0.658 & 26.35 & 23.40 & 19.56 & 18.62 & 0.687 & 27.34 \\
Qwen-3.8-Flash-Next~\cite{tongyi}
 & 27.83 & 0.544 & 0.753 & 28.31 & 25.38 & 21.04 & 24.07 & 0.679 & 21.50 \\
DeepSeek-4.1-Flash~\cite{xu2026deepseek_flash}
 & 12.34 & 0.740 & 0.933 & 10.45 & 12.74 & 12.59 & 6.95 & 0.843 & 7.62 \\
GLM-5.3-Flash~\cite{zai2026glm53flash}
 & 7.51 & 0.860 & 0.936 & 4.50 & 7.12 & 6.65 & 7.10 & 0.923 & 6.55 \\
\midrule
\rowcolor{gray!10}
\multicolumn{10}{c}{\textit{Specialized VLMs (OCR-3.0)}} \\
\textbf{Unlimited-OCR 2 (Ours)}
 & \textbf{77.06} & \textbf{0.104} & \textbf{0.123} & \textbf{87.93} & \textbf{74.81} & \textbf{71.11} & \textbf{71.79} & \textbf{0.147} & \textbf{48.20} \\

\bottomrule
\end{tabular}
}
\vspace{-0.5cm}
\end{table*}

To comprehensively evaluate Unlimited OCR 2 on document parsing across varying lengths, we conduct experiments on two challenging multi-page benchmarks: MPDocBench \cite{zhou2026mpdocbench} (up to 30 pages) and our proposed MarathonDocBench (ranging from 100 to 300 pages). Following the official evaluation protocol of MPDocBench, we report Edit Distance for textual content and reading order recognition, TEDS for table structure recognition, CDM for mathematical formula recognition, and F1 score for figure region detection. To further evaluate cross-page consistency, we include T-TextEdit and T-TEDS to assess text and table parsing across page boundaries, respectively. An overall score aggregated across these fine-grained structural categories is also computed to summarize overall parsing performance.

\begin{table}[h]
\footnotesize
\caption{Comparison on OmniDocBench v1.6. All models in the table are end-to-end VLM-based architectures.}
\vspace{-0.2cm}

\label{tab:omnidoc}
\centering
\setlength{\tabcolsep}{1.6pt}
{
\begin{tabular}{l|c|c|ccccc}
\toprule[.9pt]
{\textbf{Model}} &Size & \textbf{Overall} $\uparrow$ & \textbf{Text}$^{\text{Edit}}$ $\downarrow$  &  \textbf{Formula}$^{\text{CDM}}$ $\uparrow$ & \textbf{Table}$^{\text{TEDs}}$ $\uparrow$ & \textbf{Table}$^{\text{TEDS-S}}$ $\uparrow$ & \textbf{ReadOrder}$^{\text{Edit}}$ $\downarrow$  \\  
\midrule  
HunyuanOCR~\cite{team2025hunyuanocr}&1B & 89.95  & 0.088  & 87.68  & 91.01  & 92.23 &  0.171 \\ 
DeepSeek-OCR 2~\cite{wei2026deepseek}&3B-A0.5B & 90.25  & 0.050  & 91.84  & 83.89  & 87.75 &  0.144 \\ 
dots.ocr~\cite{li2025dots}&3B & 90.77 & 0.048 & 89.95 & 87.18 & 90.58 & 0.138 \\ 
FireRed-OCR~\cite{wu2026firered}&2B & 93.26 & 0.037  & 95.44  & 88.04  & 91.06 &  0.131 \\ 
ABot-OCR~\cite{jiang2026abot} & 2B & 93.30 & 0.037 & 94.86 & 88.69 &  91.87 & 0.137 \\
Logics-Parsing-v2~\cite{chen2025logics}&4B & 93.33  & 0.041  & 95.65  & 88.42  & 91.98 &  0.137 \\ 
Qianfan-OCR~\cite{dong2026qianfan}&4B & 93.90  & 0.040  & 95.08  & 90.53  & 93.31 &  0.130 \\ 
OvisOCR2~\cite{lu2026ovisocr2} & 0.8B  & 96.47 & 0.027 & 97.49 & 94.58 & 96.98 & 0.112\\
TeleOCR~\cite{cai2026teleocr} & 1.2B & 96.91 & 0.027 & 96.60 & 96.82 & 98.18 & 0.118\\
\midrule 
Unlimited-OCR~\cite{yin2026unlimited} &3B-A0.5B & 93.92 & 0.042 & 95.79 & 90.16 & 93.32 & 0.129 \\ 
Unlimited-OCR 2 &3B-A0.5B & 94.02 & 0.038 & 96.29 & 89.60 & 92.70 & 0.128 \\ 

\bottomrule[.9pt]
\end{tabular}
}
\end{table}

\subsection{Main Results}
\label{sec:main_results}
\vspace{-0.2cm}

As visually summarized in Figure~\ref{fig:scaling} and further detailed in Table~\ref{tab:mpd} and Table~\ref{tab:marathon}, Unlimited OCR 2 takes a significant step toward the OCR 3.0 era, advancing document parsing along three key dimensions. First, it establishes the practical feasibility of hundred-page end-to-end parsing, achieving a score of 77.06 on MarathonDocBench and demonstrating its ability to handle documents at a scale beyond conventional multi-page settings. Second, it maintains robust parsing performance as document length increases. Through end-to-end modeling with SR-SWA, Unlimited OCR 2 filters redundant visual tokens while preserving relevant cross-page information, achieving the highest Overall scores and state-of-the-art performance on both cross-page text and table metrics, T-Text and T-Table, respectively, from dozens of pages on MPDocBench to hundreds of pages on MarathonDocBench. Third, it establishes a new state-of-the-art in long-document parsing, outperforming leading closed-source models and top-tier open-source MLLMs across these benchmarks. Furthermore, Unlimited OCR 2 retains strong single-page recognition capabilities, achieving an Overall score of 94.02\% on OmniDocBench in Table~\ref{tab:omnidoc}, confirming that its sparse attention design preserves fine-grained local recognition.

\begin{table}[t]
\centering
\footnotesize
\caption{Feasibility evaluation of the Unlimited-Horizon Decoding Paradigm. Sequential chunk processing  reduces peak KV cache memory at $B=1$ and improves end-to-end throughput at $B=16$, enabling hundred-page parsing under constrained GPU memory.}
\vspace{-0.2cm}
\label{tab:split_feasibility}
\setlength{\tabcolsep}{8pt}
\renewcommand{\arraystretch}{1.2}

\begin{tabular}{l c c c}
\toprule
\textbf{Method} & \textbf{Peak KV Memory ($B=1$)} & \textbf{TPS ($B=16$)} \\
\midrule
Standard Decoding  & 47.00\,GB & 360 \\
Unlimited-Horizon Decoding   & \textbf{1.77\,GB} & \textbf{1315} \\
\bottomrule
\end{tabular}
\vspace{-0.2cm}
\end{table}

\subsection{Scaling with Document Length}

\subsubsection{Feasibility of Unlimited-horizon Parsing}
\vspace{-0.1cm}
Benefiting from SR-SWA and the Unlimited-Horizon Decoding Paradigm, Unlimited OCR 2 enables efficient parsing of extremely long documents under constrained GPU memory. We evaluate its feasibility on the hundred-page MarathonDocBench benchmark by measuring peak KV cache memory and end-to-end throughput, as reported in Table~\ref{tab:split_feasibility}. Compared with standard decoding, page-level chunking with boundary-page KV cache reuse reduces peak KV cache memory from 47.00 GB to 1.77 GB under single-sequence inference ($B=1$). Meanwhile, end-to-end throughput increases from 360 to 1,315 TPS under concurrent inference with $B=16$. These results demonstrate that sequential chunk processing with KV cache reuse substantially reduces memory requirements while improving throughput, making hundred-page document parsing more practical under limited GPU memory.
\begin{figure}[h]
	\centering
    \includegraphics[width=1\linewidth]{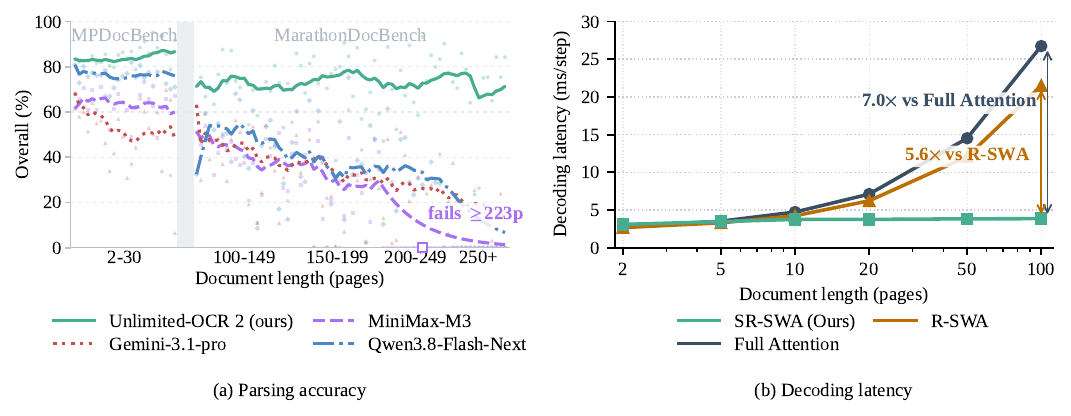}
    \vspace{-0.9cm}
	\caption{Performance and efficiency comparison across varying document lengths. Unlimited-OCR 2 demonstrates (a) robust parsing accuracy and (b) near-constant decoding latency (ms/step) as document length scaling.}
    \vspace{-0.5cm}
	\label{fig:scaling}
\end{figure}

\subsubsection{Length-Invariant Accuracy}

\vspace{-0.1cm}

To verify whether extending the document horizon compromises parsing fidelity, we evaluate the performance trajectories of Unlimited OCR 2 across continuously increasing document lengths. As illustrated in Figure~\ref{fig:scaling} (a), frontier Vision-Language Models (VLMs) typically experience a noticeable performance degradation as the sequence length grows, primarily caused by attention diffusion and context dilution over ultra-long horizons. In contrast, Unlimited OCR 2 exhibits remarkable length-invariant accuracy. By leveraging SR-SWA to dynamically isolate and retrieve only the most relevant visual tokens, our model maintains consistently high parsing precision from short sequences up to extreme horizons of 300 pages.

\subsubsection{Constant Per-Step Decoding Latency}
\label{sec:constant_latency}
\vspace{-0.1cm}

Figure~\ref{fig:scaling} (b) examines how decoding efficiency scales with document length. All measurements are conducted in Gundam mode with a batch size of 3 and a fixed generation length of 1,024 tokens per page. As the document length grows, Full Attention incurs rapidly growing per-step decoding latency because each query must attend to a linearly expanding visual context. While R-SWA mitigates this overhead by bounding local window, its remote visual scope still scales with sequence length, leading to noticeable latency accumulation over ultra-long horizons. In contrast, SR-SWA strictly bounds the effective per-query attention scope, achieving near-constant per-step decoding latency from short contexts up to hundred-page documents. Consequently, its latency and throughput advantage over Full Attention and naive windowed baselines becomes increasingly pronounced as document length scales, confirming the computational viability of our paradigm for real-world long-document parsing.

 \vspace{-0.2cm}

\section{Limitation and Future Work}
\vspace{-0.2cm}
Although Unlimited OCR 2 enables efficient parsing of hundred-page documents, its training data remains limited in diversity, and further data refinement will improve its robustness and generalization. More importantly, the current sparse-attention mechanism does not preserve global document information, limiting its understanding of document-level structures such as heading hierarchies. Future work may introduce architectures such as linear attention to retain global information with constant per-token computational cost, further improving document-level structure understanding.

 \vspace{-0.2cm}
\section{Conclusion}
\vspace{-0.2cm}

\label{sec:conclusion}

In this technical report, we presented Unlimited OCR 2, a next-generation architecture tailored for unlimited-horizon document parsing. Specifically, Unlimited OCR 2 introduces Sparse Reference Sliding Window Attention (SR-SWA) to break the long-context visual computation bottleneck, guaranteeing constant per-token computational complexity throughout decoding. Additionally, integrated with an optimized page-split decoding paradigm, the model significantly mitigates GPU memory consumption, enabling robust unlimited-horizon inference on resource-constrained hardware. Extensive evaluations and analysis confirm that Unlimited OCR 2 maintains superior parsing accuracy alongside exceptional efficiency across all scales, thereby pioneering the OCR-3.0 era of unlimited-horizon document understanding.


\section{Author List}
\noindent \textbf{Core Contributors:} Shu Wei, Jingjing Wu, Qunyi Xie

\noindent \textbf{Contributors:} Lingshu Zhang, Xiangcheng Du, Yannian Fu, Mengjie Wang, Zuyi Wang, Zhichao Liu, Yangliu Xu, Youyang Yin, Huanhuan Liu, YY, Yi Yang


\bibliography{main}

\end{CJK*}
\end{document}